\documentclass{article}

\usepackage{PRIMEarxiv}

\usepackage[utf8]{inputenc} 
\usepackage[T1]{fontenc}    
\usepackage{setspace}

\usepackage[utf8]{inputenc} 
\usepackage[authoryear,round]{natbib}
\usepackage{lineno}
\usepackage{graphicx}%
\usepackage{multirow}%
\usepackage{amsmath,amssymb,amsfonts}%
\usepackage{amsthm}%
\usepackage[title]{appendix}%
\usepackage{xcolor}%
\usepackage{lmodern}
\usepackage{textcomp}%
\usepackage{manyfoot}%
\usepackage{booktabs}%
\usepackage{bbm}
\usepackage{algorithm}%
\usepackage{algorithmicx}%
\usepackage{algpseudocode}%
\usepackage{dsfont}
\usepackage{hyperref}

\usepackage{listings}%
\usepackage{tabularx}%
\usepackage{xspace}

\usepackage{siunitx}
\usepackage{graphicx}
\usepackage{subcaption}

\newcommand{\vct}[1]{\ensuremath{\boldsymbol{#1}}}

\newcommand{\myparagraph}[1]{\smallskip \noindent \textbf{#1}}
\newcommand{\ie}{{i.e.}\xspace}
\newcommand{\eg}{{e.g.}\xspace}

\usepackage{url}
\newcommand{\iouname}{Detection Alignment\xspace}
\newcommand{\da}{\ensuremath{DA}\xspace}

\begin{document}
  
\title{Evaluating Line-level Localization Ability of Learning-based Code Vulnerability Detection Models}

\author{
  Marco Pintore\\ 
  \texttt{marco.pintore@unica.it}\\
  University of Cagliari, Italy \\
  \And
  Giorgio Piras\\
  \texttt{giorgio.piras@unica.it}\\
  University of Cagliari, Italy \\
  \And
  Angelo Sotgiu\\
  \texttt{angelo.sotgiu@unica.it}\\
  University of Cagliari, Italy \\
  CINI, Italy\\
  \And
  Maura Pintor\\ 
  \texttt{maura.pintor@unica.it}\\
  University of Cagliari, Italy \\
  CINI, Italy\\
  \And
  Battista Biggio\\
  \texttt{battista.biggio@unica.it}\\
  University of Cagliari, Italy\\
  CINI, Italy\\
}
\maketitle

\begin{abstract}
   To address the extremely concerning problem of software vulnerability, system security is often entrusted to Machine Learning (ML) algorithms. 
Despite their now established detection capabilities, such models are limited by design to flagging the entire input source code function as vulnerable, rather than precisely localizing the concerned code lines. 
However, the detection granularity is crucial to support human operators during software development, ensuring that such predictions reflect the true code semantics to help debug, evaluate, and fix the detected vulnerabilities. 
To address this issue, recent work made progress toward improving the detector's localization ability, thus narrowing down the vulnerability detection ``window'' and providing more fine-grained predictions. 
Such approaches, however, implicitly disregard the presence of spurious correlations and biases in the data, which often predominantly influence the performance of ML algorithms. 
In this work, we investigate how detectors comply with this requirement by proposing an explainability-based evaluation procedure. 
Our approach, defined as \textit{Detection Alignment} (DA), quantifies the agreement between the input source code lines that most influence the prediction and the actual localization of the vulnerability as per the ground truth. 
Through DA, which is model-agnostic and adaptable to different detection tasks, not limited to our use case, we analyze multiple learning-based vulnerability detectors and datasets. 
As a result, we show how the predictions of such models are consistently biased by non-vulnerable lines, ultimately highlighting the high impact of biases and spurious correlations. The code is available at~\url{https://github.com/pralab/vuln-localization-eval}. 
\end{abstract}


\section{Introduction}
\begin{figure}[t]
    \centering
    \includegraphics[width=\linewidth]{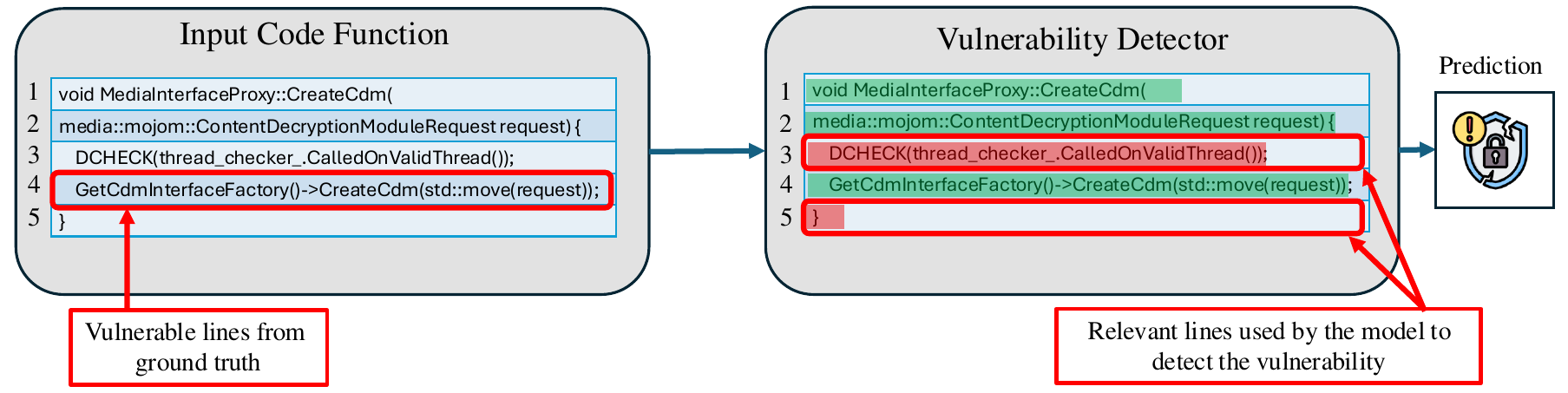}
    \caption{Ground truth vulnerable lines against the lines indicated by the detector. We show on a sample from the BigVul dataset~\citep{fan2020bigvul}, correctly predicted as vulnerable by a finetuned CodeBERT detector from~\cite{fu2022linevul}, how the localized vulnerable lines driving the prediction (3,5) are not aligned with the ground truth vulnerable line (4).}
    \label{fig:graphical_abs}
\end{figure}

From the security perspective, the last decades' fast-paced development of software has been characterized by source code vulnerabilities getting exploited by attackers. 
Such software security issues, in practice, have been represented by a growing number of threats, such as privilege escalation and remote code execution. 
The situation may be worsened by the rising use of AI-based automated code generators: while these tools accelerate development, they can introduce vulnerabilities learned from their training data~\citep{pearce2025asleep}.
To prevent the occurrence of these issues, practitioners often resort to Static Application Security Testing (SAST), which is used during software development to identify potential vulnerabilities, thus enriching the development process by providing immediate feedback. 
Conversely, to ensure correct code functioning, Dynamic Application Security Testing (DAST) is often employed to analyze the code at runtime and identify potential vulnerabilities tied to the software's dynamic behavior. 
The effectiveness of both approaches, however, is mostly constrained to detecting well-known vulnerabilities, as they typically fail to detect \textit{never-before-seen} vulnerabilities.  
Such limitation has led modern detection approaches to base their functioning on learning techniques, which can leverage large datasets of known benign and malicious code to build a predictive model for detecting new vulnerabilities~\citep{chakraborty2020deep}.
Following the success of Machine Learning (ML) techniques across different domains, a relevant portion of source code vulnerability detection techniques has leveraged such models and notably improved state-of-the-art performances~\citep{li2018vuldeepecker, zhou2019devign}. 
More recently, vulnerability detection has also been extended to transformer-based architectures~\citep{lu2021codexglue, fu2022linevul}, bringing further improvement in the detection performance. However, a now broad literature recognizes how the performance of learning-based models, which are predominantly trained in simulated or unrealistic scenarios, is often determined by the presence of spurious correlation or biases in the data, rather than generalizable patterns~\citep{chakraborty2020deep}. 
Besides limiting the real-world applicability of the models, this issue additionally emphasizes the potential of explainability techniques for studying on what evidence these models ground their decisions. In fact, it is of high relevance to ward off learning-based code vulnerability detectors to predict the presence of a vulnerability based on spurious correlations, such as less relevant syntactic or stylistic patterns not necessarily related to the actual vulnerable code portions~\citep{arp2022dos}.
These misaligned patterns can, in fact, potentially reveal that the model is not learning semantically meaningful patterns relevant to the task, but instead it is exploiting artifacts present in the data.
This becomes particularly evident when models are evaluated on unseen datasets that don't contain these artifacts and fail to generalize, highlighting the need for more robust and interpretable learning approaches.
In this regard, despite the growing number of research papers proposing code vulnerability detectors and focusing on improving their localization ability~\citep{li2021vulnerability, fu2022linevul}, it still remains unclear, from a debugging perspective, how well their detection aligns with the ground truth and neglects the influence of non-causal patterns. 
We represent this scenario in~\autoref{fig:graphical_abs}, where a real code sample from~\cite{fan2020bigvul} is correctly classified as vulnerable by a CodeBERT model~\citep{feng2020codebert, lu2021codexglue}. Nevertheless, the prediction relies on non-vulnerable lines, leading the detector to incorrectly localize the vulnerability. 

In this work, we tackle this problem by proposing an explainability-based evaluation procedure that quantifies this agreement. Our approach, which we define as Detection Alignment (\da), leverages the output of explainability techniques to assess which code lines mostly contributed to the vulnerability prediction from the ML model. 
Based on a careful design of two sets, \da quantifies how well the prediction aligns with the ground truth via Jaccard's index. 
Through our experiments, which we apply to state-of-the-art datasets and transformer-based detectors, we show how the \da metric highlights a significant influence of non-vulnerable code lines when predicting a given code function as vulnerable. 
We believe that our approach, besides being modular and adaptable to different applications outside of source code analysis, can serve as a stepping stone to improve the performance of future learning-based detectors and enhance the decision support for human operators.

\section{Background}
We describe in this section the essential concepts needed to understand the proposed method. 
As we base our Detection Alignment (\da) approach on source code vulnerability detectors, we first provide the formulation of a general learning-based detector in \autoref{sect:bg_detectors}.
While our method is general enough to cover any type of vulnerability-detection model, we present a specialization of \da on transformer-based models, and provide the main architecture notions in \autoref{sect:bg_transformers}, accordingly.
We then discuss, in \autoref{sect:bg_explainability}, the explainability concepts for the overall approach comprehension. 

\subsection{Learning-based Source Code Vulnerability Detection}\label{sect:bg_detectors}
Owing to their well-known capability of analyzing and classifying wide volumes of data, Machine Learning (ML) techniques have been dedicated over the years to the task of vulnerability identification, leveraging on Graph Neural Networks~\citep{zhou2019devign, li2021vulnerability} or Recurrent Neural Networks~\citep{li2018vuldeepecker}. 
More recently, this learning task has also been extended to transformer-based architectures~\citep{fu2022linevul}.
We provide here a general formulation that covers models predicting a label to indicate the presence of a vulnerability.

\myparagraph{General Formulation.}
Let $\vct z \in \mathcal{Z}$ be an input sample containing source code. 
The goal of a general ML model $f$ is to map $\vct z \in \mathcal{Z}$ to the label space $\mathcal{Y} = \{0, 1\}$.
Specifically, we assign the class $1$ to samples in which vulnerabilities are found, and $0$ to non-vulnerable samples.
However, as the provided source code is in a human-readable text-based format, the model is not able to process the text ``as is". 
Therefore, to convert the input into numerical data, the code is preprocessed through a feature-mapping function $\phi: \mathcal{Z} \mapsto \mathcal{X}$.
Such mapping transforms the source code file $\vct z$ to a $d$-dimensional feature vector $\vct x \in \mathbb{R}^d$.
From the newly obtained representation, the model $f: \mathcal{X} \mapsto \mathcal{Y}$ can output a prediction for the input sample $\vct z$, obtained as $\hat{y} = f(\phi(\vct z); \vct \theta)$, where $\vct \theta$ are the parameters learned by the model and the predicted label $\hat{y}$ indicates whether the input code file is predicted as vulnerable or not.  We will refer to single code samples $\vct z$ as functions, which in turn will be described by a set of $L$ code lines.

\subsection{Transformer-based Detectors}\label{sect:bg_transformers}
While our \da approach can be applied to any learning-based model, we focus our application on transformer architectures~\citep{vaswani17_nips} dedicated to the task of source code vulnerability detection, which we describe here accordingly.

\myparagraph{Tokenization Mapping.} 
The transformer preprocessing starts by converting the source code files $\vct z$, consisting of raw text, into a $d$-dimensional input viable for the model. To perform such conversion, the common procedure involves specifying a mapping function $\phi$ known as tokenization. 
The goal of a tokenizer, in the context of source code detection, is to adapt the input code to the model while preserving syntactic and structural code properties.
Most typically, tokenization is performed by (1) normalizing the input text; (2) splitting the input text into units (words, subwords, characters, bytes); (3) mapping tokens to unique integer indexes according to a precomputed vocabulary; (4) padding, truncation, and special token handling to reduce variability and allow batch processing.
The tokenization strategy can be performed at different levels of granularity, depending on the representation level that is required for the application. 
While different tokenization strategies have been proposed, we focus here on the \textit{Byte Pair Encoding} (BPE)~\citep{sennrich_bpe} sub-word tokenization approach, which has been predominantly used by transformer-based detectors.
BPE operates by first normalizing the input and then breaking it into individual characters. Then, frequently co-occurring character pairs are merged to form subwords, thus obtaining more expressive tokens while avoiding huge vocabulary sizes. To also improve efficiency on novel out-of-vocabulary words, BPE finally splits unseen words into the largest known subword units.

\myparagraph{Attention mechanism.}
Traditional text models such as Recurrent Neural Networks (RNNs)~\citep{hochreiter_lstm} process text sequentially, and often fail to capture long-range dependencies, which are widespread in source code.
Transformer architectures, on the other hand, leverage the attention mechanism to weigh the importance of different tokens in a sentence dynamically.
Specifically, transformers leverage a set of multi-head attention used to compute attention scores for every token in a sequence relating to all other tokens.
Given a head $h$ and a $d$-dimensional input embedding $\vct x$, self-attention operates by transforming each input token into three Query ($\vct Q_h$), Key ($\vct K_h$), and Value ($\vct V_h$) vectors as follows:
\begin{equation}
    \vct Q_h = \vct x \vct W^Q_h, \quad \vct K_h = \vct x \vct W^K_h, \quad \vct V_h = \vct x \vct W^V_h
\end{equation}
where $\vct W^Q_h, \vct W^K_h, \vct W^V_h \in \mathbb{R}^{d \times d_k}$ are trainable parameter matrices, and $d_k$ is the attention subspace dimension. 
Then, the attention mechanism involves obtaining a probability distribution on the dot-product between query and key, scaled by the subspace dimension $d_k$, obtaining an attention weight matrix as follows: 
\begin{equation}
    \vct A_h = \text{softmax} \left( \frac{\vct Q_h \vct K_h^T}{\sqrt{d_k}} \right)
\end{equation}
where the elements $A_{i,j}$ of the attention matrix indicate how much attention token $x_i$ puts on $x_j$. 
Finally, the output of the attention head is computed as the product between the attention weights and value matrix as $\vct H_h = \text{AttentionHead}=\vct A_h \vct V_h$.
Therefore, transformers rely on multiple heads to learn different aspects of token relationships. 
This, among others, is the case of BERT~\citep{devlin2019bert}, an encoder-only architecture relying on 12 layers composed of 12 heads respectively, adapted to code vulnerability detection via its CodeBERT~\citep{feng2020codebert} version.

\subsection{Explainable AI methods}\label{sect:bg_explainability}
Explaining a decision from ML models is useful to debug and understand what the model learns from the data.
A typical subdivision of explainability techniques involves distinguishing between \textit{local} and \textit{global} methods. 
While local explanations provide a description of specific model decisions, global ones provide a broader understanding of the model's internal representations. 
We focus here on local methods, which, for instance, can explain which tokens are more relevant and thus contribute the most to the prediction of a source code segment as vulnerable. 
In detail, we first consider raw attention scores, which can be used to understand how input tokens attend to each other and identify the most relevant ones. However, as raw attention-based explainability methods can often be unreliable, we also consider more advanced transformer-specific approaches~\citep{achtibat2024attnlrp}, as well as architecture-agnostic explanations computed through integrated gradients (IG)~\citep{sundararajan2017axiomatic}.

\myparagraph{Attention-based explainability.}
Given a transformer's encoder layer $m$, the goal of an attention-based explainability method is to determine the contribution of each token to the final prediction. Hence, considering an input sequence $\vct x$, to compute the relevance of an input sample $\vct x_i$ we are required to sum the attention scores over the different layers' heads $h$ assigned to each token $\vct x_j$ when attending to the token $\vct x_i$. A detailed implementation of our attention-based relevance computation is shown in \autoref{sect:relevance_score}.

\myparagraph{Layer-wise Relevance Propagation for Transformers.} Despite being largely applied, it has been shown that attention alone is often not sufficient to fully capture the model behavior. We thus consider a more advanced approach, Attention-aware Layer-wise Relevance Propagation (AttnLRP)~\citep{achtibat2024attnlrp}, that produces neuron-level relevance scores by rule-based backpropagation. Each network neuron is modeled as a function node that is individually decomposed, and the relevance values are distributed from the model output layer to its prior network neurons, one layer at a time. We refer to the paper for the detailed formulations.

\myparagraph{Integrated Gradients.}
While the above-described explainability measures are clearly restricted to the transformer architecture, our \da approach can also be implemented with architecture-agnostic explainability methods. 
To show this, in our work, we also consider the Integrated Gradients (IG) explainability method~\citep{sundararajan2017axiomatic}. Considering a baseline input $\vct x'$ (e.g., a sequence of padding tokens) and the actual input $\vct x$, IG measures how each input token contributes to model output by integrating gradients along a path from $\vct x'$ to $\vct x$ as follows:
\begin{equation}\label{eq:bg_ig}
    IG_i = (x_i - x_i') \int_{\alpha=0}^{1} \frac{\partial f(\mathbf{x'} + \alpha (\mathbf{x} - \mathbf{x'}))}{\partial x_i} d\alpha
\end{equation}
where $\alpha$ is a scaling parameter interpolating between baseline and input tokens. The gradient $\frac{\partial f}{\partial \vct x_i}$ measures how sensitive the model output $f$ is to changes in $\vct x_i$, effectively capturing the relevance of the input token based on how much it changes the model's decision compared to a ``neutral" baseline token. In practice, this IG formulation is approximated, for instance, through variants of Riemann or Gauss-Legendre quadrature. 
We will now show, in~\autoref{sect:da}, how we leveraged the presented explainability methods to compute three different token relevance and how, from such measure, we obtained the entire line relevance and the \da final metric.

\section{Detection Alignment}\label{sect:da}
This section presents the Detection Alignment (\da) evaluation procedure. We subdivide our approach into two specific steps: relevance score computation (\autoref{sect:relevance_score}), where we start from the tokens and compute the relevance of each line with respect to the model's output prediction; and detection alignment measurement (\autoref{sect:da_measure}), where we start from each line's relevance score and measure the overall detection alignment with respect to each input sample's ground truth.

\begin{figure}[t]
    \centering
    \includegraphics[width=\linewidth]{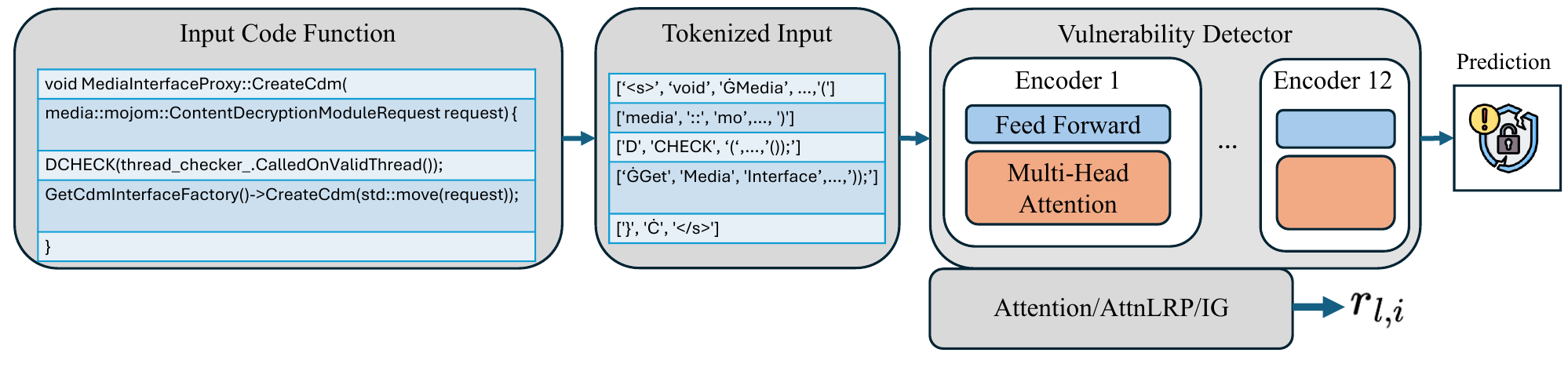}
    \caption{A code vulnerability detector. Given an input code function, we tokenize the input and use, in this use case, a transformer-based vulnerability detector to predict the presence of a vulnerability in the source code. We then compute, with an explainability-based approach (via attention scores, AttnLRP or integrated gradients), a relevance score $r_{l,i}$ for each $i$-th token of the $l$-th line.}
    \label{fig:detector}
\end{figure}
\subsection{Relevance Score Computation}\label{sect:relevance_score}  
Our \da approach starts from a pretrained code vulnerability detection model $f$, which, given a vulnerable input code function $\vct z$ (represented by $L$ code lines and mapped to a sample $\vct x$), outputs a binary label indicating whether the function is vulnerable or not. We provide an overview of the vulnerability detection model in~\autoref{fig:detector}.
Given the prediction of the vulnerable code function $\vct z$, \da aims to quantify how well it aligns with the sample's ground truth by overlapping the relevant code lines that drove the prediction with the actual ground truth vulnerable lines. We describe in the following paragraphs how, starting from tokens, we finally compute code line relevance by adopting three different explainability-based measures. 
For all these measures, however, the \da procedure starts from the tokenization step, where the input $z$, represented by $L$ lines, gets converted into a vector of tokens. 
Therefore, for each single line $l$, tokenization defines a set of $N_l$ tokens $\vct t_l = [t_{l,0}, t_{l,1}, ..., t_{l,N_l-1}]$. 
Then, considering the $i$-th token of the $l$-th line $t_{l,i}$, we compute the \textit{token relevance} using one of the three explainability-based approaches.

\myparagraph{Attention-based Relevance.}
Considering an embedded input token $t_{l,i}$, a transformer model computes a matrix of attention scores $\vct A^{(m)} \in \mathbb{R}^{H \times J \times J}$ at each encoder layer $m$, where $H$ is the number of attention heads and $J$ is the total sequence length (i.e., the total number of tokens in the input sequence $\vct z$). We compute the token-level relevance by aggregating the attention scores on a single specific layer $m$ over each of the $H$ heads as follows:
\begin{equation}\label{eq:attention_based_relevance}
    r_{l,i} = \sum^H_{h=1}\sum^J_{j=1} \vct A^{(m)}_{h,i,j}
\end{equation}
where $r_{l,i}$ is the relevance score of the $i$-th token of line $l$, $h$ the attention head, and $j$ represent each of the other input tokens. Therefore, $\vct A^{(m)}_{h,i,j}$ represents the attention score assigned to token $r_{l,i}$ in layer $m$ by the $H$ heads when attending to the $j$-th token. 
Therefore, the higher the aggregated attention score $r_{l,i}$, the more that token $t_{l,i}$ contributed to the final prediction. 
However, while attention scores provide a straightforward way to interpret model decisions, they can often have limited reliability in our source code detection task due to biases in the code syntactic structure or positional dependencies~\citep{achtibat2024attnlrp}. 
In turn, we do not limit our relevance computation to such a procedure but also extend it to the more advanced AttnLRP-based and architecture-agnostic IG-based computation as follows.

\myparagraph{AttnLRP-based relevance.} As the AttnLRP method is already able to provide a relevance score for each network's neuron, it is sufficient to take the produced score $r_{l,i}$ with respect to each input token.

\myparagraph{Integrated Gradients-based Relevance.}
The third approach we use to compute token relevance instead, leverages layer integrated gradients (LIG)~\citep{sundararajan2017axiomatic}. 
In detail, with respect to the base IG formulation from~\autoref{eq:bg_ig}, we compute the gradients with respect to the input embeddings, effectively capturing the model attribution, approximating the integral with the Gauss-Legendre quadrature rule as follows: 
\begin{equation}\label{eq:lig_approx}
r_{l,i} = \sum_{k=1}^{s} w_k \cdot \frac{\partial F(x' + \alpha_k (x - x'))}{\partial x_i}
\end{equation}
where the values $w_k$ weight the sum on a set of $s$ steps, and $\alpha_k \in [0,1]$ defines the interpolation between the baseline token $\vct x'$ to the actual input $\vct x$.

\myparagraph{Line-based Relevance Score.} We subsequently aggregate the token-based relevance scores $r_{l,i}$ into line-based relevance $r_l$ as follows: 
\begin{equation}\label{eq:rl}
r_l = \sum_{i=1}^{N_l} r_{l, i}.
\end{equation}
where $N_l$ represents the number of tokens for line $l$. After repeating this process on the entire function, a line relevance score can be assigned to each line, resulting in a vector $\vct r = [r_0, r_1, ..., r_{L-1}]$. Since different explainability methods can produce relevance scores on substantially different scales, we rescale each line's relevance in $[0, 1]$ to ensure consistency and proportionality across different methods.

\begin{figure}[t]
    \centering
    \includegraphics[width=\linewidth]{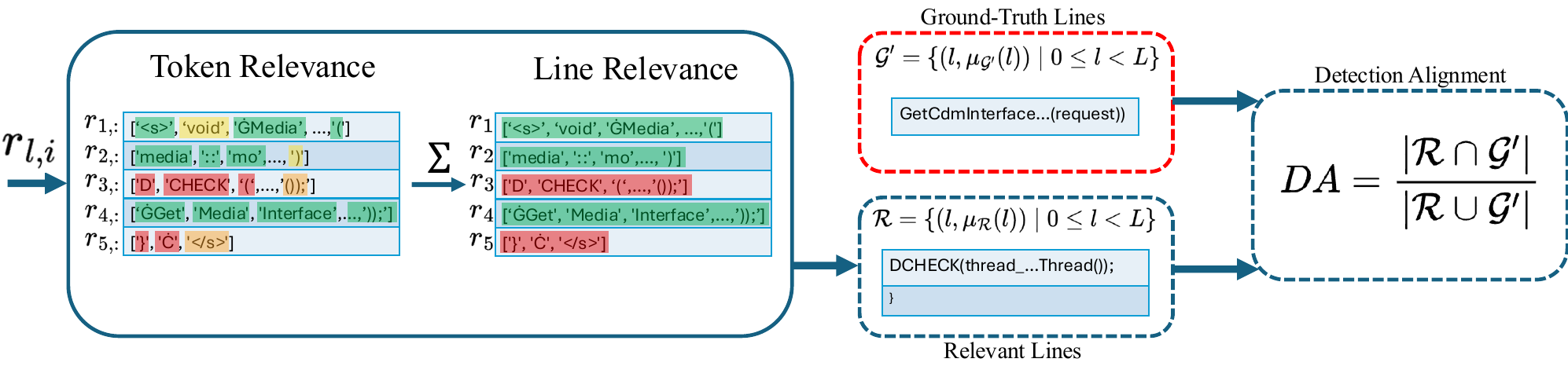}
    \caption{A \da approach overview. We start from the token relevance scores $r_{l,:}$, which we aggregate into line relevance scores $r_l$. Then, we design the two sets: $\mathcal{G}^\prime$, derived from the ground truth $\mathcal{G}$; and $\mathcal{R}$, derived from the line relevance scores $r_l$. We finally compute \da through the Jaccard's index computed between the two sets. }
    \label{fig:da}
\end{figure}
\subsection{Detection Alignment Measurement}\label{sect:da_measure}
After obtaining a set of relevance scores for each line $\vct r$, as we preliminary show in~\autoref{fig:da}, we can now effectively compute the alignment between the lines that drove the model prediction and the ground truth for the vulnerable input sample $\vct z$, represented as a set $\mathcal{G} \subseteq \{ l \mid 0 \leq l < L\}$, which contains the indices of the actual vulnerable code lines. 
To compute \da, we take inspiration from fuzzy-set theory~\citep{PETKOVIC2021107849}, which enables the concept of partial membership in a set through a membership degree function $\mu \in [0,1]$. 
Therefore, in more practical terms, $\mu$ indicates the degree of belonging of the element to the fuzzy set. 

We implement such a concept in a first fuzzy-set $\mathcal{R}$, where each element is represented by the line index $l$ along with its line relevance score $r_l$, representing the elements' membership degree $\mu_{\mathcal{R}}(l) = r_l$.
This amounts to designing a first set  $\mathcal{R} = \{ (l, \mu_{\mathcal{R}}(l)) \mid 0 \leq l < L\}$. 
We then apply the same logic to create a second fuzzy-set $\mathcal{G}^\prime$, which is the result of converting $\mathcal{G}$ into $\mathcal{G}^\prime = \{(l, \mu_{\mathcal{G^\prime}}(l)) \mid 0 \leq l < L \}$, where $\mu_{\mathcal{G^\prime}}(l) = \mathds{1}_{\mathcal{G}}(l)$ is the membership function of $\mathcal{G}^\prime$, which simply assigns ones when the line is vulnerable, as per the ground truth. After creating the two sets, we compute the detection alignment \da through the Jaccard's index (i.e., intersection over union) between the fuzzy sets $\mathcal{R}$ and $\mathcal{G}^\prime$ as follows: 
\begin{equation}\label{eq:da}
    \da = \frac{\lvert \mathcal{R} \cap \mathcal{G}^\prime \rvert}{\lvert \mathcal{R} \cup \mathcal{G}^\prime \rvert} =
    \frac{\sum_{l} \min(\mu_{\mathcal{R}}(l), \mu_{\mathcal{G^\prime}}(l))} {\sum_{l} \max(\mu_{\mathcal{R}}(l), \mu_{\mathcal{G^\prime}}(l))},
\end{equation}
where $0 \leq \da \leq 1$, with $1$ as optimal value.
Such \da computation, leveraging the Jaccard's index while extending on fuzzy sets, allows for capturing the extent to which both prediction and ground truth align on line importance (intersection) while accounting for all lines that are considered relevant by one of ground truth or prediction (union).
To measure the detection alignment, \autoref{eq:da} is applied only on samples that are predicted as vulnerable by the model, and assign $\da=0$ to vulnerable samples predicted as benign. Overall, our \da metric increases when the model assigns high relevance to ground truth lines and decreases when, in contrast, the model assigns high relevance to the lines that do not correspond to the vulnerability. 
Abstracting from a single code function $z$, we average the metric over multiple input samples $z \in \mathcal{Z}$ and obtain a global \da estimate.

We point out that the \da can also be generalized to diverse input representations other than tokens (\eg, graphs), with the only requirement of being able to map back the attributions produced with respect to the input representation onto the relative source code portions. Furthermore, the metric can be applied with different levels of granularity than code lines (\eg, group of tokens corresponding to source code entities), assuming to have a coherent ground truth.

\section{Experiments}
We present here the experiments conducted to evaluate our Detection Alignment (\da) metric. We start from a set of state-of-the-art vulnerability detectors and datasets, which we describe in \autoref{sect:exp_setup}. Then, in \autoref{sect:results}, we show and discuss the results obtained by our \da approach along with the other standard evaluation metrics, highlighting the misalignment of the predictions from such vulnerability detectors.

\subsection{Experimental Setup}\label{sect:exp_setup}
We validate our approach on 3 datasets and 3 transformer-based vulnerability detectors. In the next paragraphs, we first discuss the details concerning both models and datasets. Then we provide an overview of the adopted experimental setup, concluding by presenting the \da computation settings.

\myparagraph{Models.}
We evaluate 3 transformer-based vulnerability detection models: LineVul~\citep{fu2022linevul}, CodeBERT~\citep{lu2021codexglue}, and  CodeT5-Small~\citep{wang2021codet5}. 
We describe here the three detection models:  
\begin{itemize}
    \item \textbf{CodeBERT} is a pretrained language model for programming languages~\citep{feng2020codebert}. However, in our specific use case, we refer to the version finetuned on a vulnerability detection task~\citep{lu2021codexglue}. Like other BERT architectures~\citep{devlin2019bert}, it is based on a total of 12 encoder-only layers.
    
    \item \textbf{LineVul} is a model designed for fine-grained vulnerability prediction~\citep{fu2022linevul}. It is built on the pretrained CodeBERT language model, which follows the BERT architecture~\citep{devlin2019bert} with 12 encoder-only layers. 
    
    \item \textbf{CodeT5-Small} is a unified encoder-decoder transformer model devised for multiple tasks. In our case, we use the model in its defect detection finetuned version~\citep{wang2021codet5}, based on the T5 architecture consisting of 6 encoder layers and 6 decoder layers~\citep{raffael2020t5}.
\end{itemize}

Among the used models, LineVul proposes to analyze the attention mechanism to improve the granularity of the detection, thus aiming to localize the vulnerable code lines. CodeBERT and CodeT5 are instead limited to a coarse-grained vulnerability prediction, thus simply flagging the input source code function as vulnerable or not.

\myparagraph{Datasets.} We use 3 different datasets to validate the effectiveness of our \da metric. In detail, we finetune each of the three models on BigVul~\citep{fan2020bigvul}, Devign~\citep{zhou2019devign}, and Primevul~\citep{ding2025primevul}.
The datasets can be described as follows:

\begin{itemize}
    \item \textbf{BigVul} is built by collecting a large set of C/C++ code from 348 open-source GitHub repositories. The authors crawled the public Common Vulnerabilities and Exposures (CVE) database and related source code repositories~\citep{fan2020bigvul}.
    We split the dataset into benign and vulnerable samples following the same approach adopted in~\cite{fu2022linevul}, ending up with a dataset consisting of 188.636 functions, of which 177.736 are benign and 10.900 are vulnerable.
    
    \item \textbf{Devign} is constructed by employing security-related keywords to filter out GitHub commits~\citep{zhou2019devign}.
    In addition to this automated dataset construction, the vulnerable commits have been further annotated by human operators. However, as pointed out by~\cite{ding2025primevul}, the annotations from Devign are highly inaccurate. We split the dataset following~\citep{lu2021codexglue}, which results in an overall of 27.318 functions, of which 14.858 are benign and 12.460 vulnerable. 
    
    \item \textbf{PrimeVul} is a recently proposed benchmark that aims to overcome biases in existing datasets \citep{ding2025primevul}. It is built by merging security-related commits from several previously proposed datasets, including BigVul, while excluding Devign after discovering that many of its commits were unrelated to security issues. The authors applied several steps to address common biases like the presence of code duplicates, incorrect labeling, and chronological inconsistencies between train and test data that cause data leakage. It consists of 235,768 samples, with 228,800 non-vulnerable and 6,968 vulnerable functions.

\end{itemize}
We summarize the main characteristics of these datasets in \autoref{tab:datasets}. We base our choice of the datasets on their popularity among other work on vulnerability detection and their availability~\citep{ding2025primevul}. 
Although we finetune the three models on each of these datasets, obtaining a total of 9 different models, to test each detector our approach requires a line-level ground truth $\mathcal{G}$ (see~\autoref{sect:da_measure}), which indicates which are the vulnerable lines within a vulnerable input function. 
In turn, we test each of the 9 models on the BigVul test set~\citep{fan2020bigvul}, composed of $18864$ samples, of which $1005$ are vulnerable. This dataset provides a field \texttt{vul\_func\_with\_fix} (where, for each vulnerable function, the flawed lines are flagged), from which we extracted the ground truth set $\mathcal{G}$ following the same methodology as in~\citep{fu2022linevul}.

\begin{table}[t]
\centering
\caption{Summary of datasets used for finetuning and evaluation.}
\label{tab:datasets}
\begin{tabular}{lrrrr}
\toprule
\textbf{Dataset} & \textbf{Language} & \textbf{\# Functions} & \textbf{\# Vulnerable} & \textbf{\# Benign} \\
\midrule
BigVul    & C/C++ & 188,636 & 10,900  & 177,736 \\
Devign    & C/C++ & 27,318  & 12,460  & 14,858  \\
PrimeVul  & C/C++ & 235,768 & 6,968   & 228,800 \\
\bottomrule
\end{tabular}
\end{table}

\myparagraph{Finetuning Settings.} 
We finetune each of the three models with the three different datasets, adding up to a total of 9 different models. In detail, for all the model-dataset pairs, we finetune using a learning rate $\gamma = 2 \times 10^{-5}$, a number of epochs equal to $10$, and a fixed number of processed tokens equal to $512$, i.e., each input function is allowed to be represented by a maximum of $510$ tokens, and special tokens marking the start and the end of the sequence are added. If the length of the tokenized source code is higher, it is truncated; otherwise, padding tokens are used to fill the input sequence until it reaches the required length. 
Among the $1055$ samples labeled as vulnerable used to compute the \da metric, $426$ samples are too long to be entirely represented in the input sequence and thus are truncated to $510$ tokens.
During the finetuning procedure, we save the models with the highest F1 scores based on validation.
We summarize the setup and model-dataset combinations in \autoref{tab:models}.
\begin{table}[t]
\centering
\caption{Summary of models used in the experiments.}
\label{tab:models}
\begin{tabular}{lcccc}
\toprule
\textbf{Model} & \textbf{Architecture} & \textbf{Granularity} & \textbf{Finetuned On} & \textbf{Tested On} \\
\midrule
CodeBERT       & Encoder-only   & Function-level & BigVul, Devign, PrimeVul & BigVul \\
LineVul        & Encoder-only   & Line-level     & BigVul, Devign, PrimeVul & BigVul \\
CodeT5-Small   & Encoder-Decoder & Function-level & BigVul, Devign, PrimeVul & BigVul \\
\bottomrule
\end{tabular}
\end{table}

\myparagraph{Metrics.} 
We compute the relevance scores on each of the 9 models following the approaches described in~\autoref{sect:relevance_score}. In detail, we first use the attention-based relevance computation on two specific encoder layers out of $M$: the first ($m=1$) and the last ($m=M$), for which we provide a motivation in~\autoref{sect:results}. Based on these two different attention-based relevance measures, we define a \da-$A^{(1)}$ and \da-$A^{(M)}$ detection alignment metric. Then, we apply the AttnLRP method, obtaining a \da-$ALRP$ detection alignment metric.
Finally, we use the architecture-agnostic integrated gradient approach to compute relevance, from which we derive a third \da-$IG$ detection alignment metric. 
These four \da metrics are computed on each of the 9 model/dataset pairs, on which we also compute F1-score values to provide a complete overview of the detectors' performances, which would otherwise be limited. We summarize all the employed metrics in~\autoref{tab:metrics}.
\begin{table}[t]
\centering
\caption{Evaluation metrics used to assess model performance and alignment.}
\label{tab:metrics}
\begin{tabular}{ll}
\toprule
\textbf{Metric} & \textbf{Description} \\
\midrule
F1-score        & Standard classification performance metric on function-level predictions. \\
\da-$A^{(1)}$   & Detection Alignment based on attention weights from the first encoder layer. \\
\da-$A^{(M)}$   & Detection Alignment based on attention weights from the last encoder layer. \\
\da-$ALRP$   & Detection Alignment using Attention Layer-wise Relevance Propagation. \\
\da-$IG$        & Detection Alignment computed via Integrated Gradients (architecture-agnostic). \\
\bottomrule
\end{tabular}
\end{table}

\begin{table}[t]
\centering
\caption{F1 score and \da computed on the BigVul~\citep{fan2020bigvul} test set with the attention from the first attention layer (\da{-}$A^{(1)}$), the last layer $m$ (\da{-}$A^{(M)}$), through AttnLRP (\da{-}$ALRP$) and Integrated Gradients (\da{-}$IG$). We denote with - the models for which we could not obtain acceptable performances (F1$<0.01$).}
\resizebox{\textwidth}{!}{%
  \begin{tabular}{l l c c c c c c}
 \textbf{Training set} & \textbf{Model} & \textbf{F1} & \boldmath{$\da\text{-}A^{(1)}$} & \boldmath{$\da\text{-}A^{(M)}$} &
 \boldmath{$\da\text{-\textbf{ALRP}}$} &
 \boldmath{$\da\text{-\textbf{IG}}$}\\
 \toprule
 \multirow{3}{*}{BigVul} & CodeBERT &  0.40 & 0.1241 & 0.0938 & 0.1140 & 0.1142\\
 & LineVul & 0.91 & 0.1208 & 0.0992 & 0.0850 & 0.0958 \\
 & CodeT5 & 0.93 & 0.1224 & 0.1118 & 0.0782 & 0.0777 \\
 \midrule
 \multirow{3}{*}{Devign} & CodeBERT &  0.13 & 0.1181 & 0.1012 & 0.1093 & 0.1047\\
 & LineVul & 0.11 & 0.1278 & 0.0906 & 0.1160 & 0.1116 \\
 & CodeT5 & 0.10 & 0.1343 & 0.0922 & 0.1314 & 0.1381 \\
 \midrule
 \multirow{3}{*}{PrimeVul} & CodeBERT & 0.17 & $4.33 \times 10^{-5}$ & $4.21 \times 10^{-5}$ & $2.14 \times 10^{-5}$ & 0.0001 \\
 & LineVul & {-} & {-} & {-} & {-} & {-} \\
 & CodeT5 & 0.29 & 0.0045 & 0.0046 & 0.0107 & 0.0062 
 \\
 \bottomrule
 \end{tabular}
}
\label{tab:results} 
\end{table}
\subsection{Results}\label{sect:results}
Based on the model and dataset pairs previously described, we create a set of 9 models that we test on the BigVul dataset. Overall, the results obtained from our \da approach highlight, in \autoref{tab:results}, how the line-level localization of such detectors is consistently misaligned with the ground truth. This implies that, despite obtaining acceptable F1-score values, the predictions provided by these models are mostly driven by code lines that are not vulnerable, suggesting the presence of spurious correlations or biases when correctly flagging a vulnerable code function. 
For completeness, we computed the metric also considering the absolute values of the relevance scores, \ie $r_l = \sum_{i=1}^{N_l} \lvert r_{l, i} \rvert$, obtaining results very close of the ones without the absolute values.

\myparagraph{Relevance Level Comparison.} 
We compute four \da metrics based on four different relevance computations. We choose to compute \da on the first layer, \da{-}$A^{(1)}$, inspired by (and to enable comparison with) the line-level localization approach from~\cite{fu2022linevul}, which explicitly uses the attention on the first layer to indicate the vulnerable lines of code. 
Then, following the rationale that growing in layers the models can provide more class-specific representations in different tasks and architectures~\citep{zeiler2014visualizing}, we additionally use attention-based relevance to compute \da{-}$A^{(M)}$ on the last encoder layer $M$ and compare with the first layer \da metric. 
Overall, we find that the \da{-}$A^{(M)}$ metric is consistently lower compared to \da{-}$A^{(1)}$, suggesting that the first encoder layers can capture more meaningful patterns based on the attention scores. More broadly, this result shows that the first and last layers simply focus on different kinds of correlations, with a lower alignment on the last layer. 
However, independently of the layer, we find a consistently low \da metric, suggesting how, in general, such detectors fail in flagging the input code functions as vulnerable based on the actual vulnerable lines. 
A similar conclusion can also be drawn for the AttnLRP and integrated gradients based metrics \da{-}$ALRP$ and \da{-}$IG$, which in some specific cases also suggests worse alignment than the attention-based metrics and assumes higher relevance owing to its improved reliability, as mentioned in~\autoref{sect:bg_explainability}.
Comparing the F1-score and the various \da metrics through Spearman's rank correlation, we observe a moderate negative correlation ($\rho\approx-0.61$) across all metrics. However, this trend lacks statistical significance ($p\text{-value} > 0.05$) and is further limited by the small sample size, which prevents us from drawing any strong scientific conclusions. 

\myparagraph{Model \& Dataset Level Comparison.}  
Independent of the model-dataset combination, we constantly find a low \da. In general, through our setup, we train and test models with the same dataset on the first three BigVul models and additionally analyze the generalization capabilities by training the models on Devign and PrimeVul while still testing on BigVul. 
Interestingly, by analyzing the F1-score, we find that the performance on the Devign dataset is consistently low for all models. 
We believe that this condition also occurs due to the low label accuracy affecting Devign~\citep{zhou2019devign}. 
In fact, previous work has shown that the label accuracy for vulnerable functions on Devign is limited to 24\%, which inevitably impairs the performance of the trained detectors. 
However, for both Devign and PrimeVul, the low F1-score inevitably highlights a general lack of detectors to generalize to different datasets.  
Also, we find the LineVul model trained on PrimeVul unable to reach appropriate performances, having an almost negligible F1 score. 
Models trained on BigVul instead clearly hold higher F1. 
Notably, though, the \da obtained by such models is extremely low, which further indicates how, despite having an extremely high F1 score, this metric alone is not entirely descriptive of the detectors' capabilities. In fact, most of the correct vulnerability flags given to the input code functions are actually due to non-vulnerable lines.
Comparing across the different models instead, we likewise find constantly lower \da, without any highly relevant deviance from such a trend.

\begin{figure}[t]
    \centering
    \begin{subfigure}{0.45\textwidth}
        \centering
        \includegraphics[width=\textwidth]{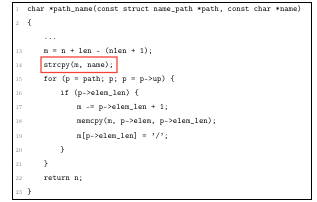}
        \caption{}
        \label{fig:gt_strcpy}
    \end{subfigure}
    \hfill
    \begin{subfigure}{0.45\textwidth}
        \centering
        \includegraphics[width=\textwidth]{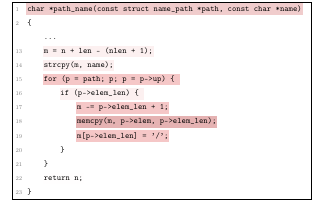}
        \caption{}
        \label{fig:vuln_strcpy}
    \end{subfigure}
    \caption{A vulnerable sample from the BigVul dataset. Figure (a) depicts the ground truth vulnerable line (14), while Figure (b) shows the relevant lines (\da{-}$A^{(1)}$) as conceived by a CodeBERT model trained on the Devign dataset. The detector mostly assigns high relevance to non-vulnerable code lines (1, 15, 17, 18, and 19).}
    \label{fig:sample_strcpy}
\end{figure}

\begin{figure}[t]
    \centering
    \begin{subfigure}{0.45\textwidth}
        \centering
        \includegraphics[width=\textwidth]{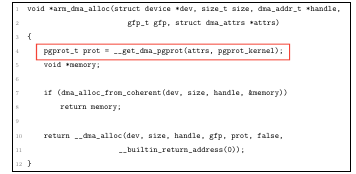}
        \caption{}
        \label{fig:gt_alloc}
    \end{subfigure}
    \hfill
    \begin{subfigure}{0.45\textwidth}
        \centering
        \includegraphics[width=\textwidth]{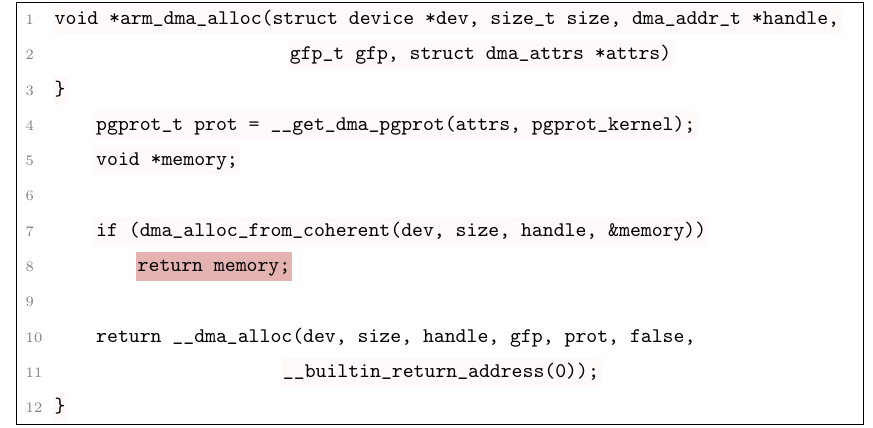}
        \caption{}
        \label{fig:vuln_alloc}
    \end{subfigure}
    \caption{A vulnerable sample from the BigVul dataset. Figure (a) depicts the ground truth vulnerable line (4), while Figure (b) shows the relevant lines (\da{-}$A^{(1)}$) as conceived by a CodeBERT model trained on the Devign dataset. The detector assigns high relevance to non-vulnerable code line (8).}
    \label{fig:alloc}
\end{figure}

\myparagraph{Sample-wise Analysis.} 
To provide a more detailed analysis of the effectiveness of our approach, we show in \autoref{fig:sample_strcpy} and \autoref{fig:alloc} a qualitative analysis of two samples, considering the CodeBERT model and attention-based relevance at the first layer as the attribution method. \autoref{fig:sample_strcpy} shows a vulnerable sample potentially related to a \emph{buffer overflow} attack, localized in line 14 according to the ground truth provided by the BigVul dataset. Although the sample is correctly labeled as vulnerable by the model, the computed line relevance scores show how the prediction was mostly driven by non-vulnerable lines misaligned with the ground truth. 
In detail, as the \texttt{strcpy} function allows copying strings of arbitrary length into the memory region pointed by \texttt{m}, if the \texttt{name} length exceeds the allocated space, this can potentially lead to a buffer overflow. 
However, the model assigns high relevance to pointer arithmetic and, in particular, to low-level memory operations such as the \texttt{memcpy} function on line 18. Interestingly, among the 10,018 vulnerable samples in the Devign dataset, \texttt{strcpy} appears in only 36 samples, whereas \texttt{memcpy} occurs 766 times. This disparity suggests that the model may associate the presence of \texttt{memcpy} with vulnerability, highlighting the influence of spurious correlations learned from training data.
Coherently, the $\da{-}A^{(1)}$ of this sample is equal to $0.0421$, being $\lvert \mathcal{R} \cap \mathcal{G}^\prime \rvert = 0.3538$ and $\lvert \mathcal{R} \cup \mathcal{G}^\prime \rvert = 8.3921$. 
In contrast, the model fails to correctly predict the vulnerable sample in~\autoref{fig:alloc}. 
In this case, the ground truth in~\autoref{fig:alloc}(a) indicates the vulnerability in line 4, where an executable DMA (Direct Memory Access) mapping might lead to undesired privilege escalation. However, the resulting relevance attribution in~\autoref{fig:alloc}(b) suggests that the model's prediction is mostly led by line 8, hence implying no intersection with the ground truth and resulting in $\da{-}A^{(1)}$ of $0$ ($\lvert \mathcal{R} \cap \mathcal{G}^\prime \rvert = 0$ and $\lvert \mathcal{R} \cup \mathcal{G}^\prime \rvert = 1$).

\section{Related Work}
We describe here the most relevant related work, which we subdivide based on work focusing on the detectors' interpretability (\autoref{sect:related_interpretable}), and on studies analyzing biased decisions in vulnerability detection (\autoref{sect:related_bias}). 

\subsection{Interpretable Vulnerability Detectors}\label{sect:related_interpretable}
The majority of proposed approaches only produce a binary label indicating the presence of a vulnerability in the analyzed source code input. 
Existing literature focuses on developing models that could provide fine-grained predictions, \eg at the line level, in order to provide more meaningful feedback for human operators in charge of analyzing vulnerabilities.
VulDeeLocator
\citep{li2022vuldeelocator} uses intermediate code representations, human-defined rules, and a Bidirectional Recurrent Neural Network (BRNN) enhanced with additional layers to produce line-level outputs. Similarly, VulChecker \citep{mirsky2023vulchecker} exploits a GNN trained on program dependency graphs produced from intermediate code representation, while LineVD\citep{hin2022linevd} combines a transformer model and a GNN to achieve the same objective.
LineVul \citep{fu2022linevul} and IVDetect \citep{li2021vulnerability} leverage the attributions produced with attention scores from a transformer-based model and explainable AI algorithms from a GNN, respectively, to obtain and output the most important lines of code for each model's prediction.
First, we highlight that in our work the focus is slightly different. 
We do not focus on human interpretability, but rather on exposing the model’s reliance on spurious correlations that hinder generalization to unseen data.
In our work, we compute the \iouname on LineVul (in addition to other models providing coarse-grained outputs), as we restricted the analysis to transformer-based approaches. This gives us the possibility to test our metric with both model-specific (\ie, attention scores) and architecture-agnostic (\ie, Integrated Gradients) attribution methods. 
We note that our approach is agnostic of both considered models and attribution methods and thus can be easily applied to the other aforementioned detectors. Furthermore, for those models, the \iouname can be computed relying either on the produced line-level outputs or on an attribution method.

\subsection{Bias in Vulnerability Detectors}\label{sect:related_bias}
Several works inspected vulnerability detection models through explainable machine learning techniques, trying to determine how they produce a certain output to understand if their decisions were the result of a real capture of the meaning of the analyzed code, or if they were affected by bias and exploited shortcuts to label the data. 
\citet{chakraborty2020deep}, applied attribution methods on some input samples to extract the most important features used to classify them from 4 different detectors, showing that in most cases they focused on code segments not correlated with the vulnerability present in the code.
\citet{warnecke2020evaluating} showed how an RNN-based vulnerability detector \citep{li2018vuldeepecker} relies on artifacts present in the dataset that are irrelevant to the discovery of vulnerabilities in the code. The same problem was later resumed and analyzed in detail in \citet{arp2022dos}. \citet{sotgiu2022explainability} extended these sample-wise analyses to the dataset level, considering also transformer-base detectors.
Although all these works share the use of explainability methods to investigate how much the decisions of the models are taken based on an understanding of the code and on the real location of the vulnerabilities, all of them require the intervention of a human operator who inspects the outputs produced, to evaluate and quantify this phenomenon. Contrarily, we provide a metric that allows us to have an immediate vision and a quantification of the problem.
\citet{imgrund2023broken} presented a method to quantify how much detectors are influenced by some categories of artifacts in the code (e.g., identifier names, coding styles, etc.), but they do not consider how much a model is able to focus on the relevant parts of the code.
\citet{warnecke2020evaluating} also proposed metrics to evaluate, under multiple perspectives, the quality of explanations produced by different algorithms on models applied to security-related tasks. \citet{ganz2021explaining} extended this analysis by specifically considering GNNs for vulnerability detection. Following the same line of work, \citet{ganz2023hunting} proposed other criteria to compare different explainability techniques by specifically focusing on vulnerability detection and considering different detectors. More generally, a body of research~\citep{lukas2021when, ARRAS202214, GUIDOTTI2021103428, Zhou_Booth_Ribeiro_Shah_2022} investigated approaches to evaluate the quality of xAI methods, leveraging ground truth on synthetic or real-world data. These works are orthogonal to ours, as they do not focus on models but rather on techniques to evaluate the quality and utility of attributions. 
These techniques can, however, be used to select the best attribution methods to use to compute our metric, helping to disentangle whether observed discrepancies with the ground truth truly reflect model misalignment and not shortcomings in the explanation methods themselves.

\section{Conclusions}
Despite ML-based source code vulnerability detectors reporting promising results, their performance is often overestimated, especially when they rely on learning feature correlations that are not relevant with respect to the presence of a vulnerability. Recent work has explored this phenomenon through the lens of empirical qualitative analysis, but having a complete overview of it requires an inspection by a human expert.

In this work, we propose the \iouname (\da) metric, which quantifies how much a model is able to focus on the input portions (\eg, code lines) that are actually related to the considered problem when it predicts a sample as vulnerable. We implement the metric by extracting the most relevant lines influencing the classifier's decision with explainable AI algorithms and comparing them with a ground truth representing the lines associated with the vulnerabilities. We test the metric on the BigVul dataset, considering a combination of three transformer-based detectors trained on three different datasets, showing that all of them struggle to recognize the really important parts for the presence of a vulnerability in the code. This is also confirmed by the lack of generalization capabilities across different datasets. We believe \da can provide a fast and easy-to-interpret estimate of the vulnerability detectors' trustworthiness, which can help debug and improve models during their development.
Moreover, the metric design allows practitioners to adapt it to a wide range of diverse settings (models, datasets, programming languages) and extend it to several tasks beyond vulnerability detection (code quality assessment, bug finding), and domains (natural language processing, and with slight adaptations, even computer vision).

In future work, we plan to extend the experimental evaluation in order to consider different attribution methods and detectors, possibly able to handle larger inputs and avoid truncating the input functions, which represent a limitation of current approaches.
Beyond evaluation, a key direction is to integrate the \da metric directly into the training process. In particular, we propose leveraging \da as a regularization signal that encourages the model to focus its predictive attention on semantically relevant code regions.

\section*{Acknowledgements}
This work has been partly supported by the EU-funded Horizon Europe projects ELSA (GA no. 101070617) and Sec4AI4Sec (GA no. 101120393); and by projects SERICS (PE00000014) and FAIR (PE00000013, CUP: J23C24000090007) under the MUR NRRP funded by the European Union - NextGenerationEU.

\bibliographystyle{plainnat}  
\bibliography{main_preprint}  
\end{document}